\documentclass[aps,pra,twocolumn,superscriptaddress,a4paper,english]{revtex4-1}

\usepackage{float}
\usepackage[english]{babel}
\usepackage{physics}
\usepackage{graphicx}
\usepackage{dcolumn}
\usepackage{bm}
\usepackage{hyperref}
\usepackage[mathlines]{lineno}
\usepackage{amssymb}
\usepackage{tcolorbox} 
\usepackage{enumerate} 

\usepackage{xcolor}
\hypersetup{
	colorlinks,
	linkcolor={blue!80!black},
	citecolor={blue!80!black},
	urlcolor={blue!80!black}
}
\usepackage[caption=false]{subfig}
\makeatletter
\newcommand{\colorcaption}[2][]{%
	\begingroup%
	\renewcommand{\@caption@fignum@sep}{ (color online). }%
	\caption[#1]{#2}%
	\endgroup%
}
\makeatother

\begin{document}
	\title{Entanglement-guided architectures of machine learning by quantum tensor network}
	\author{Yuhan Liu}
	\affiliation{Department of Physics, Sun Yat-sen University,  Guangzhou 510275, China}
	\author{Xiao Zhang}
	\affiliation{Department of Physics, Sun Yat-sen University,  Guangzhou 510275, China}
	\author{Maciej Lewenstein}
	\affiliation{ICFO - Institut de Ciencies Fotoniques, The Barcelona Institute of Science and Technology, Av. Carl Friedrich Gauss 3, 08860 Castelldefels (Barcelona), Spain}
	\affiliation{ICREA, Pg. Llu\'is Companys 23, 08010 Barcelona, Spain}
	\author{Shi-Ju Ran}
	\email[Corresponding author. Email: ]{shi-ju.ran@icfo.eu}
	\affiliation{ICFO - Institut de Ciencies Fotoniques, The Barcelona Institute of Science and Technology, Av. Carl Friedrich Gauss 3, 08860 Castelldefels (Barcelona), Spain}
	
	\date{\today}
	
	\begin{abstract}
		It is a fundamental, but still elusive question whether the schemes based on quantum mechanics, in particular on quantum entanglement, can be used for classical information processing and machine learning. Even partial answer to this question would bring important insights to both fields of machine learning and quantum mechanics. In this work, we implement simple numerical experiments, related to pattern/images classification, in which we represent the classifiers by many-qubit quantum states written in the matrix product states (MPS). Classical machine learning algorithm is applied to these quantum states to learn the classical data. We explicitly show how quantum entanglement (i.e., single-site and bipartite entanglement) can emerge in such represented images. Entanglement characterizes here the importance of data, and such information are practically used to guide the architecture of MPS, and improve the efficiency. The number of needed qubits can be reduced to less than $1/10$ of the original number, which is within the access of the state-of-the-art quantum computers. We expect such numerical experiments could open new paths in charactering classical machine learning algorithms, and at the same time shed lights on the generic quantum simulations/computations of machine learning tasks.
	\end{abstract}
	\maketitle
	\section{Introduction}
	
Classical information processing mainly deals with pattern recognition and classification. The classical patterns in question may correspond to images, temporal sound sequences, finance data, and so on. During the last thirty years of developments of the quantum information science, there were many attempts  to generalize classical information processing to the quantum world, for instance by proposing quantum perceptrons and quantum neural networks (e.g., see some early works \cite{lewenstein1994quantum,QNN1,QNN2} and a review \cite{QNNrev1}), quantum finance (e.g., \cite{B07fqbook}), quantum game theory \cite{Eisert2000,Neil2001,DLXS+2002Qgame}, to name a few. More recently, there were successful proposals to use quantum mechanics to enhance learning processes by introducing quantum gates/circuits, or quantum computers \cite{DTB16MLqenhance,DLWT17MLqenhance,L17MLqcircuit,MSW17MLenhance,HGSSG17MLmera,WLqml}.

Conversely, there were various attempts to apply methods of quantum information theory to classical information processing tasks, for instance by mapping classical images to quantum mechanical states.  In 2000, Hao et al. \cite{Hao00} developed a different representation technique for long DNA sequences, obtaining mathematical objects similar to many-body wave-function. In 2005 Latorre \cite{Latorre05} developed independently a mapping between bitmap images and many-body wavefunctions which has a similar philosophy, and applied quantum information techniques in order to develop an image compression algorithm. Although the compression rate was not competitive with standard JPEG, the insight provided by the mapping was of high value \cite{Le11}. A crucial insight for this work was the idea that Latorre's mapping might be inverted, in order to obtain bitmap images out of many-body wavefunctions. In fact, in Ref. \cite{qubism} developed a reverse idea, and mapped quantum many-body states to images.

Such an interdisciplinary field was recently strongly motivated, due to the exciting achievements in the so-called ``quantum technologies'' (see some general introductions in, e.g., \cite{Qtech00,Qtech01,Qtech1,Qtech2}). Thanks to the successes in quantum simulations/computations, including the D-Wave \cite{Dwave} and the quantum computers by Google and others (``Quantum Supremacy'') \cite{Google1,Google2}, it becomes unprecedentedly urgent and important to explore the utilizations of quantum computations to solve machine learning tasks.

Particularly, a considerable progress has been made in the field merging quantum many-body physics and quantum machine learning \cite{BWPR+17QML} based on tensor network (TN) \cite{SS16TNML,HWFWZ17MPSML,LRWP+17MLTN,S17MLTN,PV17MLTNlanguage,LYCS17EntML}. TN provides an powerful mathematical structure that can efficiently represent many-body states, operators, and quantum circuits, even though the dimension of the Hilbert (vector) space suffers an exponential growth with the size of the system \cite{CV09TNSRev,O14TNSRev,O14TNadvRev,RTPCSL17TNrev}. Paradigm examples include matrix product states (MPS) \cite{O14TNSRev}, projected entangled pair states \cite{VC06PEPSArxiv,O14TNSRev}, tree TN states \cite{SDV06TTN}, or multi-scale entanglement renormalization ansatz \cite{V07EntRenor}. Recently, TN proved its great potential in the field of machine learning, providing a natural way to build the mathematical connections between quantum physics and classical information. Among others, MPS has been utilized to supervised image recognition \cite{SS16TNML} and generative modeling to learn joint probability distribution \cite{HWFWZ17MPSML}. Tree TN that has a hierarchical structure is also used to natural language modeling \cite{PV17MLTNlanguage} and image recognition \cite{LRWP+17MLTN,S17MLTN}, which is proven to be of high efficiency. The relations between the mathematical models of machine learning, e.g., Boltzmann machine and TN states, MPS and string-bond state, and deep convolutional arithmetic circuits and quantum many-body wave functions, have been investigated \cite{CCXWX17TNML,GPARC17MLTN,HM17TNML,LYCS17EntML}. 

Furthermore, it is worth mentioning that (classical) machine learning techniques have been introduced to solve physical problems. For example, it has been proposed to use neural networks to learn quantum phases of matter, and detect quantum phase transitions \cite{CM17MLphase,BAT17MLphase,BCMT17MLfermion,CCMK17MLfermion,CVK17MLphase,ZK17MLphase,HSS17MLphase,W16MLphase,BDSW+17MLphase,TT16MLphase,CT17MLphys,HDW17MLphase,PWS16MLphase,CL17NNstate,NDYI17MLquantum}. Different schemes of machine learning, including supervised, unsupervised, and reinforcement learning, are applied to systems of spins, bosons and fermions, combined with gradient methods, Monte Carlo, and so on. 

Despite these inspiring achievements, there are several pressing challenges. One of those concerns how to practically utilize quantum features or even quantum simulations/computations to process classical data \cite{PanJW,Seth2013,MLscCircuit}. With the existing methods (e.g., \cite{SS16TNML,LRWP+17MLTN,S17MLTN}), the number of the qubits is the same as the number of pixels in an image, which is too large to be realized with the current techniques of quantum computations. Anther challenge relates to the underlying relations between the properties of classical data and those of quantum states (e.g., quantum entanglement), which are still elusive.


In this work, we implement simple numerical experiments with MPS, and show how quantum entanglement can emerge from images and be used for the learning architecture. We encode sets of images consisting of pixels of a certain shade of grey, onto the many-qubit states in a Hilbert space \cite{SS16TNML}. The classifiers of the encoded images are represented as MPS's. A training algorithm based on Multiscale Entanglement Renormalization Ansatz (MERA) \cite{MERA,LRWP+17MLTN} is then used to optimize the MPS. We show, considering the images before and after the discrete cosine transformation (DCT), that the efficiency of such classical computation is characterized by the bipartite entanglement entropy (BEE). The MPS for classifying the images after DCT possesses much smaller BEE, meaning higher efficiency, than the MPS for the images before DCT. The single-site entanglement entropy (SEE) of the trained MPS's characterizes the importance of the local data (e.g., different pixels). This permits to discard the less important data, so that the number of the needed qubits can be largely reduced. Our simulations show that to reach the same accuracy, the number of qubits ($28 \times 28=784$ qubits originally) for classifying the images after DCT can be lowered about ten times compared with that for classifying before DCT. Furthermore, we propose to optimize the MPS architecture according to SEE, and achieve in this way higher computational efficiency smaller number of qubits without harming the accuracy. The reduced number of qubits (about $50 \sim  100$) is accessible to the current techniques of quantum computations.

\section{Review of matrix product state and training algorithm}

\begin{figure}
	\centering
	\includegraphics[width=0.9\linewidth]{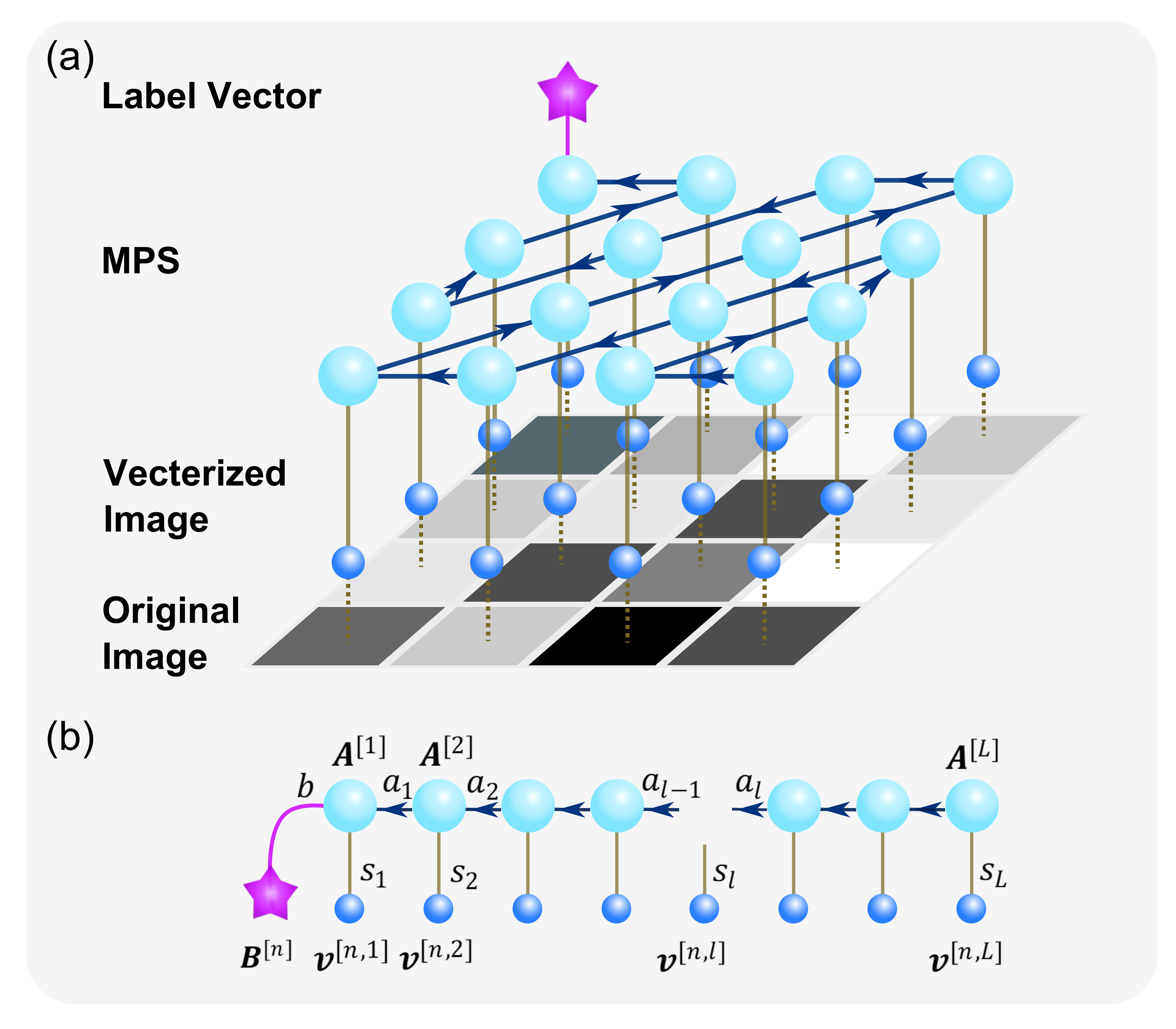}
	\caption{Illustration of (a) MPS $\mathbf{\hat{\Psi}}$ and (b) the environment tensor $\mathbf{\mathcal{E}}^{[n,l]}$. The MPS covers the 2D image in a Zigzag path. The original images (either before or after DCT) will be vectorized into many-qubit states by the feature map [Eq. (\ref{eq-featuremap})]. $\mathbf{\hat{\Psi}}$ satisfies the orthogonal condition, indicated by the arrows. $\mathbf{\mathcal{E}}^{[n,l]}$ is defined by contracting everything after taking out the tensor (blue) that is to be updated.}
	\label{fig-Env}
\end{figure}

The basic idea is after mapping the classical data into a vector (quantum Hilbert) space, quantum states (or the quantum operator formed by these states) are trained to capture different classes of the images, in order to solve specific tasks such as classifications. Since the Hilbert space is usually exponentially large when the size of the images increases, TN (MPS in this work) are to implement the calculations efficiently by classical computers.

\subsection{Feature from data to quantum space}

Such a TN machine leaning contains two key ingredients. One is the feature map \cite{S17MLTN} that encodes each input image to a product state of many qubits. Each pixel (say, the $l$-th pixel $\theta_{n,l}$ of the $n$-th image) is transformed to a qubit given by $d$-dimensional normalized vector as
\begin{equation}
v^{[n,l]}_s=\sqrt{\binom{d-1}{s-1}}[\cos(\frac{\pi }{2} \theta_{n,l})]^{d-{s}} [\sin(\frac{\pi}{2}\theta_{n,l})]^{s-1},
\label{eq-featuremap}
\end{equation}
where $s$ runs from 1 to $d$. We take $d=2$ in this work, and each qubit state satisfies $|v^{[n,l]}\rangle = v^{[n,l]}_1 |\uparrow\rangle + v^{[n,l]}_2 |\downarrow\rangle$. Then, the $n$-th image is mapped to a $L$-qubit state, which is a $d^L$-dimensional tensor product state $\prod_{l=1}^L |v^{[n,l]}\rangle$ ($L$ is the number of pixels of the image). One can see that the number of qubits equals to the number of pixels in one image. Note that in the paper, we use the bold symbols to represent tensors without explicitly writing the indexes.

\subsection{Tensor network representation and training algorithm}

\begin{figure}
	\centering
	\includegraphics[width=\linewidth]{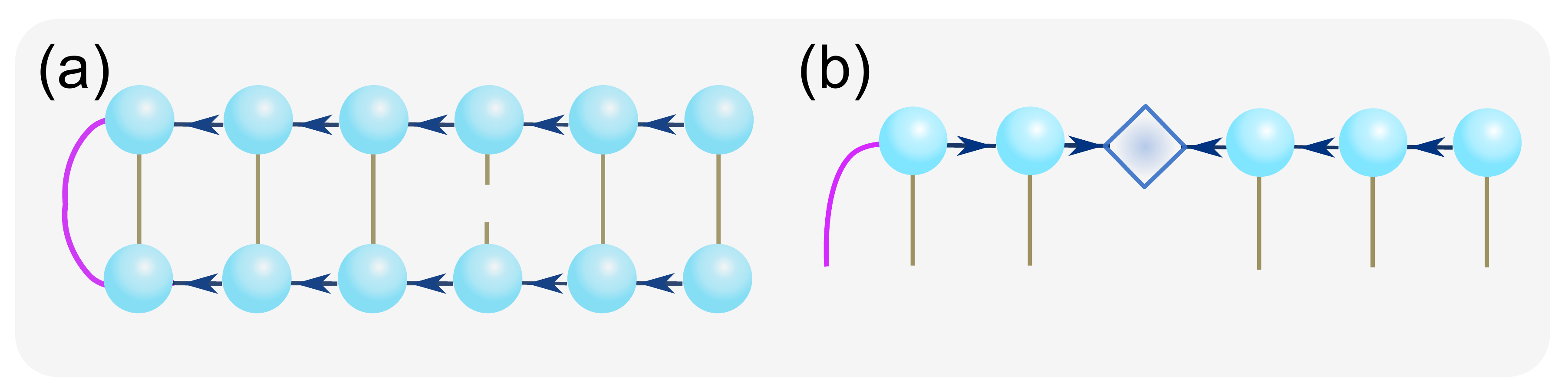}
	\caption{Computation of (a) the single-site entanglement entropy and (b) bipartition entanglement entropy.}
	\label{fig-EntDef}
\end{figure}

The second key ingredient is the TN. The output of the $n$-th image is obtained by contracting the corresponding vectors with a linear projector denoted by $\hat{\mathbf{\Psi}}$ as $|u^{[n]} \rangle = \hat{\mathbf{\Psi}} \prod_{l=1}^L |v^{[n,l]}\rangle$. Its coefficients satisfy
\begin{equation}
u^{[n]}_b = \sum_{s_1 \cdots s_L} \hat{\Psi}_{b,s_1 \cdots s_L} \prod_{l=1}^L v^{[n,l]}_{s_l}.
\label{eq-costf}
\end{equation}
$\hat{\mathbf{\Psi}}$ is actually a map from a $d^L$-dimensional to a $D$-dimensional vector space. Here, we take $\hat{\mathbf{\Psi}}$ as a unitary MPS (Fig. \ref{fig-Env}) whose coefficients satisfy
\begin{equation}
\hat{\Psi}_{b,s_1 \cdots s_L} = \sum_{a_1 \cdots a_{L-1}} A^{[1]}_{s_1ba_1} A^{[2]}_{s_2a_1 a_2} \cdots A^{[l]}_{s_la_{l-1}a_l} \cdots A^{[L]}_{s_La_{L-1}}.
\label{eq-MPS}
\end{equation}
Note the indexes $\{a\}$, which are often called virtual bonds, will be all summed over. The dimensions of the virtual bonds (denoted by $\chi$) determines the upper bound of the entanglement that can be carried by the MPS. The $d$-dimensional indexes $\{s_l\}$ are called physical bonds, which are contracted with the encoded images $\{|v^{[n,l]}\rangle\}$. The total number of parameters in the MPS increases only linearly with $L$, i.e. $O(d\chi^2L)$. For implementing the contraction between $\{|v^{[n,l]}\rangle\}$ and the MPS, one should choose a 1D path that covers the 2D image. Here, we choose a zig-zag path shown in Fig. \ref{fig-EntDef} (a), and later show that such a path can be optimized according to the SEE of the MPS.

To train the MPS, we optimize the tensors $\{\mathbf{A}^{[l]}\}$ in the MPS one by one to minimize the error of the classification. To this end, the cost function to be minimized is chosen to be the simplified negative log likelihood(NLL) $f^{CE} = -  \sum_{n}  \ln(\sum_{b} B^{[n]}_{b} u^{[n]}_{b})$, with $\mathbf{B}^{[n]}$ a $D$-dimensional vector ($D$ is the number of classes) that satisfies
\begin{eqnarray}
B^{[n]}_{b} &=&
\left\{
\begin{array}{lll}
1, \ \ \text{if the n-th image $\in$ the \textit{b}-th class} \\
0, \ \ \text{otherwise}
\end{array}
\right.
\end{eqnarray}

We use the MERA-inspired algorithm to optimize the MPS \cite{LRWP+17MLTN}, where all tensors are taken as isometries that satisfy the right orthogonal condition $\sum_{s_l a_l} A^{[l]}_{s_l,a_{l-1}a_l} A^{[l]}_{s_l,a_{l-1}'a_l} = I_{a_{l-1} a_{l-1}'}$ (for the rightmost tensor, it still satisfies this condition by considering it as a $\chi\times d\times 1$ tensor). Then the MPS in Eq. (\ref{eq-MPS}) gives a unitary projector from a $d^L$-dimension to a $D$-dimensional vector space. The tensors in the MPS can be initialized randomly, and then are optimized one by one (from right to left, for example). The key step is to calculate the (unnormalized) environment tensor $\mathbf{\mathcal{E}}^{[n,l]}$, which is defined by contracting everything after taking out the target tensor $\mathbf{A}^{[l]}$ (see Fig. \ref{fig-Env} (b) and the supplementary material for details). Then, define $\mathbf{E}^{[l]} = \sum_{n} \mathbf{\mathcal{E}}^{[n,l]} / (\sum_{bs_1 \cdots s_L} B^{[n]}_b \hat{\Psi}_{b,s_1 \cdots s_L} \prod_{l=1}^L v^{[n,l]}_{s_l})$ and use SVD as $\mathbf{E}^{[l]} = \mathbf{U} \mathbf{\Lambda} \mathbf{V}^{\text{T}}$. The tensor is updated by $\mathbf{A}^{[l]} \leftarrow \mathbf{U} \mathbf{V}^{\text{T}}$. One can see that the new tensor still satisfies the orthogonal condition. Update all tensors in this way one by one until they converge. The code can be found on GitHub \cite{github}.

\subsection{Discrete cosine transform and motivation}

In addition, we try the standard discrete cosine transformation (DCT) to transform the images in frequency space before feeding them to the MPS. The DCT is defined as
\begin{equation}
\begin{aligned}
\eta_{u,v}=&\frac{2}{M}\alpha(u)\alpha(v) \sum_{x=0}^{M-1}\sum_{y=0}^{M-1} \\ &\theta_{x,y}\cos{[\frac{(2x+1)u\pi}{2M}]}\cos{[\frac{(2y+1)v\pi}{2M}]},
\end{aligned}
\end{equation}
with $M$ the width/height of the images, $(x,y)$ the position of a pixel, and $\alpha(u)=1/\sqrt{2}$ if $u=0$, or $\alpha(u)=1$ otherwise. In our case, we have $M=28$ for the images in the MNIST dataset. Note $L=M^2$.

We propose that DCT is very helpful while choosing the path of the MPS to deal with 2D images. In the frequency space, there exists a natural 1D path for this. The zig-zag path shown in Fig. \ref{fig-EntDef} (a) is used in many standard image algorithms (e.g., JPEG). The frequency is non-increasing along the path. Note that in previous works using MPS, the 2D images are directly reshaped into 1D (i.e., $(1 \times M^2)$) images. 

Moreover, it is known from the existing image algorithms that the most important information is normally stored in the low-frequency data. It is interesting to see if the entanglement of the trained MPS reveals the same property. In this way, the number of qubits can be further reduced when defining the MPS on the zig-zag path and training after DCT transformation.

\section{Learning architecture based on quantum entanglement}

We will show below that by learning the images from the frequency space (reached by DCT), the computational cost can be largely reduced without lowering the accuracy. This is revealed by a lower BEE of the MPS, which means that smaller virtual bond dimensions are needed. More interestingly,	we propose a learning architecture based on quantum entanglement to further improves the efficiency. The architecture contains two aspects: optimizing the MPS path according to SEE, and discarding less important data according to BEE. Our work practically utilize (bipartite and single-site) quantum entanglement to design machine learning algorithms for classical data. It exhibits an explicit example of ``quantum learning architecture''. We test our proposal with MNIST dataset of handwriting digits \cite{MNIST}.




\subsection{Single-site and bipartite entanglement entropy of MPS}

Before presenting our results, let us define the single-site entanglement entropy (SEE) and bipartite entanglement entropy (BEE). The reduced density matrix $\mathbf{\hat{\rho}}^{[l]}$ of the $l$-th site, for example, is defined as
\begin{equation}	
\hat {\rho}^{[l]}_{s_ls_l'}=
\sum_{b s_1 \cdots s_{l-1} s_{l+1} \cdots s_L} \hat{\Psi}_{b s_1 \cdots s_l \cdots s_L} \hat{\Psi}_{b s_1 \cdots s_l'\cdots s_L}.
\end{equation}
Note $\mathbf{\hat{\rho}}^{[l]}$ is non-negative. The computation of $\mathbf{\hat{\rho}}^{[l]}$ with MPS is shown in Fig. \ref{fig-EntDef} (b), where one contracts everything except the indexes $s_l$ and $s_l'$. The leading computational complexity is about $O(ld\chi^3)$ by using the orthogonal condition. After normalizing $\mathbf{\hat{\rho}}^{[l]}$ by $\mathbf{\hat{\rho}}^{[l]} \leftarrow \mathbf{\hat{\rho}}^{[l]} / \text{Tr} \mathbf{\hat{\rho}}^{[l]}$, the SEE is defined as
\begin{equation}
\begin{aligned}
S_{\text{SEE}}^{[l]} =-\text{Tr}\mathbf{\hat{\rho}}^{[l]} \ln \mathbf{\hat{\rho}}^{[l]}.
\end{aligned}
\end{equation}

The BEE measured between, for example, the $l$-th and $(l+1)$-th sites is similarly defined by the reduced density matrix obtained after tracing over either half of MPS. There is another way to obtain BEE by singular value decomposition (SVD), where BEE is given by the singular values (or called Schmidt numbers). The SVD is formally written as
\begin{equation}
\begin{aligned}
\hat{\Psi}_{b s_1 \cdots s_l s_{l+1} \cdots s_L} = \sum_{aa'} X_{b s_1 \cdots s_l,a} \lambda^{[l]}_{aa'} Y_{a',s_{l+1} \cdots s_L},
\end{aligned}
\end{equation}
where the singular values are given by the non-negative diagonal matrix $\mathbf{\lambda}^{[l]}$, and $X$ and $Y$ satisfy the orthogonal conditions $\mathbf{X} \mathbf{X}^{\text{T}} = \mathbf{Y}^{\text{T}} \mathbf{Y} = \mathbf{I}$. Normalizing $\mathbf{\lambda}^{[l]}$ by $\mathbf{\lambda}^{[l]} \leftarrow \mathbf{\lambda}^{[l]} /|\mathbf{\lambda}^{[l]}|$, BEE is defined as
\begin{equation}
\begin{aligned}
S_{\text{BEE}}^{[l]} =-\sum_{a} \lambda_{aa}^{[l]2} \ln \lambda_{aa}^{[l]2}.
\end{aligned}
\end{equation}

The computation of BEE in our context is illustrated in Fig. \ref{fig-EntDef} (c). One only needs to transform the first $(l-1)$ tensors to the left orthogonal form (indicated by the arrows), then $\mathbf{\lambda}^{[l]}$ is obtained by the SVD of $\mathbf{A}^{[l]}$ as $A^{[l]}_{s_l a_{l-1} a_l} = \sum_{aa'} X_{s_la_{l-1},a} \lambda_{aa'}^{[l]} Y_{a',a_l}$. The leading computational cost is $O(ld\chi^3)$.


\begin{figure}
	\centering
	\includegraphics[angle=0,width=1\linewidth]{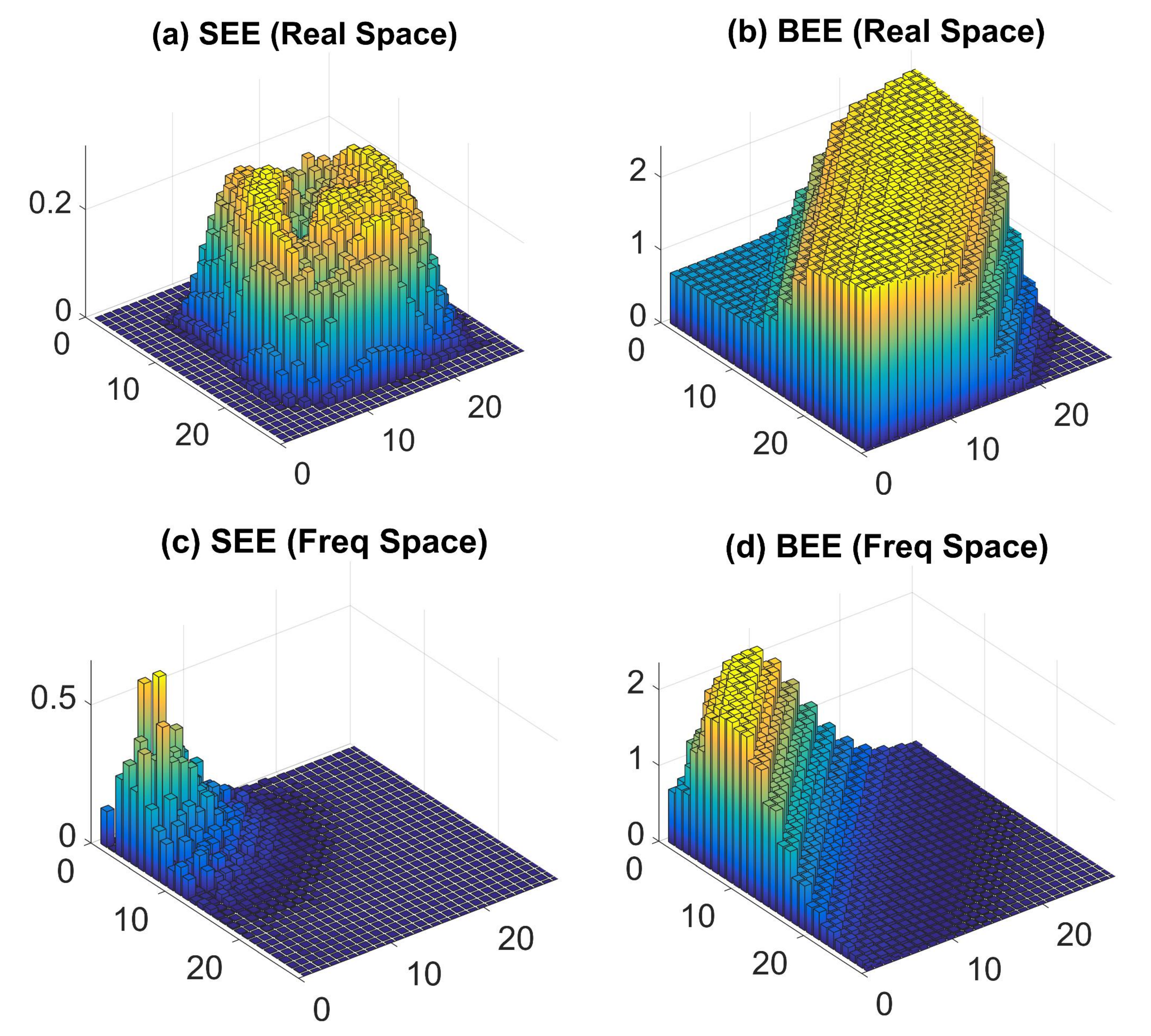}
	\caption{(a) Single-site entanglement entropy (SEE) and (b) bipartite entanglement entropy (BEE) of MPS without DCT. (c) and (d) show the SEE and BEE with DCT. We take the classification between the images ``0'' and ``2'' as an example. The virtual bond dimension is $\chi=16$, with $D=2,d=2$.}
	\label{fig-Ent2D}
\end{figure}

\begin{figure}
	\includegraphics[angle=0,width=1\linewidth]{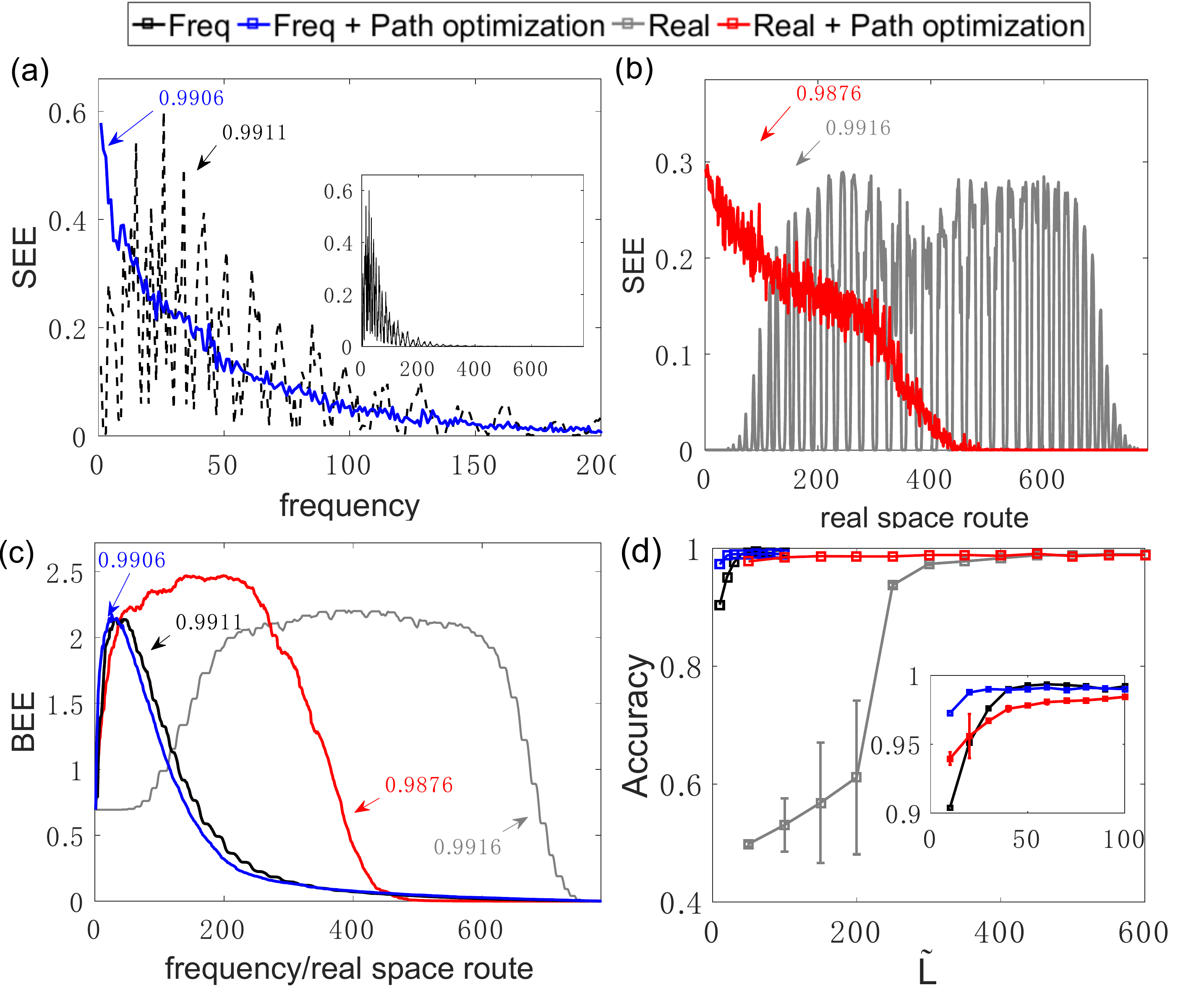}
	\caption{(a) SEE in frequency space without and with path optimization according to SEE, (b) SEE in real space, (c) BEE in real space and frequency space, without and with path optimization, and (d) accuracy on the test dataset for different MPS length $\tilde{L}$. The virtual bond dimension is $\chi=16$, with $D=2$ and $d=2$. The accuracies are also indicated. In (d), the accuracies from the real-space data suffer large fluctuations, indicated by the error bar.}
	\label{fig-reorder}
\end{figure} 

In Fig. \ref{fig-Ent2D} (and most of the paper), we take the MPS trained for classifying images ``0'' and ``2'' as an example, and show its SEE and BEE with and without the DCT. Without DCT, the data are in the real space, i.e., simply the pixels of the 2D images. The relatively large values of SEE are distributed almost all over the 2D plane. With DCT, the data are in fact the weights of different frequencies. The large values of the SEE only appears in the positions that are close to the label bond. 

SEE actually characterizes the amount of non-trivial information carried by the data. Without DCT, the important information is distributed almost all over the 2D plane. See supplementary material C for more details. With DCT, the important information to the classification problem are mainly of low frequencies. This is consistent with what is know from the well-established image algorithms, that the low-frequency data are more important. With our work, such a phenomenon is naturally justified by the values of SEE of the trained MPS. 

Meanwhile, the BEE with DCT increases in a much slower way than that without DCT. Due to the orthogonal conditions of the MPS, the information flows from the right end of the MPS to the left (label bond). Each time when the non-trivial information (indicated by a relatively large SEE) is passed through, BEE increases and finally saturates to a finite value around $\ln \chi$. While approaching to the label bond on the left end, BEE decreases to $\ln D$, giving a triangular plateau of BEE [see Fig. \ref{fig-Ent2D} (b)]. This can be understood as a ``refining'' process: while the information flows to the label bond (output), only the the information that is important to the classification will be kept. The value $\ln D$ of the BEE also indicates that the state of each virtual bond in the plateau is actually described by the two-qubit maximally entangled state.

In the MPS schemes, it is well-known that the BEE determines the needed dimensions of the corresponding virtual bond. Particularly, when the entanglement entropy vanishes to zero, it means the corresponding data is uncorrelated to others and need not be fed to the MPS. In the following, we will show that to reach the same accuracy, smaller length of MPS, meaning less qubits, are needed with DCT than without DCT. This provides an efficient scheme to discard the sites of small SEE.

\subsection{Learning architecture based on single-site entanglement entropy}

\begin{table*}[tbp]
	\centering
	\begin{tabular}{c|p{13.5cm}}
		\hline\hline
		\textit{\textbf{Step 1}} & Randomly initialize the MPS, choose a path (say, zigzag), and train it by the standard algorithm; calculate the SEE of the MPS. \\
		\hline
		\textit{\textbf{Step 2}} & Redefine the path according to the values of SEE at different sites. \\
		\hline
		\textit{\textbf{Step 3}} & Define the MPS on the new path, randomly initialize it, and train it.\\
		\hline
		\textit{\textbf{Step 4}} & Calculate the SSE: if the SSE is in an acceptable descending order, end the training; if not, go back to Step 2. \\
		\hline
		\textit{\textbf{Step 5}} & Calculate the BEE and find the $\tilde{L}$-th site where BEE equals to $0.75\ln D$. Discard the data after this site ($l>\tilde{L}$) and train the new MPS with the length of $\tilde{L}$.\\
		\hline\hline
	\end{tabular}
	\caption{Steps of the training algorithm, where the architecture of the MPS is guided by the entanglement.}
	\label{tab-steps}
\end{table*}

To minimize the BEE, we propose to rearrange the path of the MPS, so that the SEE is in a non-ascending order. The steps are listed in Table \ref{tab-steps}. After path optimization, the BEE will be lowered, meaning the computational cost will be lowered, while the accuracy remains unchanged.

\begin{table*}[tbp]
	\centering
	\begin{tabular}{ c||c|c|c|c|c|c|c|c|c } 
		\hline
		--& 1 & 2 &3&4&5&6&7&8&9 \\
		\hline\hline
		0 & 0.15(0.76) & 0.11(0.82)& 0.13(0.81)& 0.12(0.82)& 0.11(0.80)& 0.13(0.79)& 0.13(0.80)&0.12(0.81)& 0.13(0.82)  \\ 
		\hline
		1 & - & 0.13(0.76)& 0.15(0.79)& 0.15(0.75)& 0.14(0.77)& 0.15(0.77)& 0.15(0.76)& 0.15(0.76)& 0.15(0.77) \\
		\hline
		2& -& -& 0.11(0.82)& 0.11(0.81)& 0.11(0.80)& 0.11(0.79)& 0.11(0.81)& 0.11(0.81)& 0.12(0.83)\\
		\hline
		3& -& -& -& 0.12(0.81)& 0.11(0.80)& 0.13(0.80)& 0.13(0.81)& 0.13(0.81)& 0.14(0.81)\\
		\hline
		4& -& -& -& -& 0.12(0.80)& 0.12(0.80)& 0.13(0.80)& 0.12(0.80)& 0.13(0.80)\\
		\hline
		5& -& -& -& -& -& 0.13(0.80)& 0.13(0.80)& 0.12(0.80)& 0.13(0.81)\\
		\hline
		6& -& -& -& -& -& -& 0.13(0.80)& 0.13(0.80)& 0.13(0.81)\\
		\hline
		7& -& -& -& -& -& -& -& 0.13(0.80)& 0.14(0.81)\\
		\hline
		8& -&-& -& -& -& -& -& -& 0.14(0.81)\\
		\hline
		\hline
	\end{tabular}
	\caption{Complexity ratios $\xi$ [Eq. (\ref{eq-ratio})] of classifiers trained by frequency data  and by the real-space data (shown in the bracket) without path optimization.}
	\label{tab-ratio}
\end{table*}

To explain how this architecture works, let us give a simple example with a three-qubit quantum state. The wave function reads $|\psi \rangle =\left|\uparrow \uparrow \downarrow \right\rangle + \left|\downarrow \uparrow \uparrow \right\rangle$, where $\left|\uparrow\right\rangle$ and $\left|\downarrow\right\rangle$ stand for the spin-up and spin-down states, respectively. By writing the wave function into a three-site MPS, one can easily check that the two virtual bonds are both two-dimensional. The total number of parameters of this MPS is $2^2 + 2^3 + 2^2 = 16$. However, if we define the MPS after swapping the second qubit to either end of the chain, say swapping it with the third qubit, the wave function becomes $|\psi \rangle = \left|\uparrow \downarrow \uparrow \right\rangle + \left|\downarrow \uparrow \uparrow \right\rangle = (\left|\uparrow \downarrow \right\rangle + \left|\downarrow \uparrow  \right\rangle) \otimes \left| \uparrow \right\rangle$. Obviously, the virtual bonds of the MPS are two- and one-dimensional, respectively, and the total number of parameters is reduced to $2^2 + 2^2+2 = 10$. In our algorithm, the SSE will normally be in a good descending order after optimizing the path only once.

Fig. \ref{fig-reorder} (a) shows the SEE in the frequency space with and without path optimization. Without path optimization, the important data where the values of SSE are relatively large are distributed on the first 200 sites (see the inset of Fig. \ref{fig-reorder} (a)). By zooming in this range, one can see that the SSE are in a good descending order after optimizing the path. For comparison, we show in Fig.\ref{fig-reorder} (b) the SEE of the MPS trained by the real-space data with and without optimizing the path.

Fig. \ref{fig-reorder} (c) shows the BEE, which indicates the computational cost of using the MPS to solve the classification task. It is obvious that the BEE of the MPS trained by the frequency data is much smaller than that of the MPS trained by the real-space data. By path optimization, the BEE is further reduced, indicating that smaller bond dimensions are needed.

Fig. \ref{fig-reorder} (d) shows the accuracy when discarding certain less important data. We only use the first $\tilde{L}$ data of each image to train the $\tilde{L}$-site MPS. We observe that as $\tilde{L}$ increases, the accuracy trained with the frequency data rises quickly and reach the value more than $0.98$ with $\tilde{L}$ being as small as 40. For comparison, training by the real-space data obviously requires a larger number of qubits, which can be reduced significantly by optimizing the path. The reduced number is almost comparable to that with DCT. For the training after DCT, the difference between the accuracies with and without path optimization is relatively small. This is because we take $\chi=16$, where the maximal capacity of the entanglement entropy ($\ln \chi$) is much larger than the reduction of the BEE by the path optimization.


To characterize the improvement of efficiency that can be gained by discarding the less important data, we define the \textit{complexity ratio}
\begin{equation}\label{eq-ratio}
	\xi=\frac{\tilde{L}}{L}. 
\end{equation}
$\tilde{L}$ is defined by a threshold, so that the BEE is smaller than $c \ln D$ when measured after the $\tilde{L}$-th site. $c$ is a number determined by the requirement of accuracy. We take $c=0.75$. When $\xi \ll 1$, it means the data on the last $(1-\xi) L$ sites can be ignored without harming the accuracy too much. Our results show that $\xi=0.82$ when trained with real-space data without path optimization, and $\xi=0.11$ and $0.10$ using frequency data without and with path optimization, respectively. More results are given in Table \ref{tab-ratio}. We show that the trainings by the data with and without DCT lead to similar accuracies, but the efficiencies (characterized by the complexity ratios) are largely different.

	
	\section{Summary and prospects}
	
	In this work, we explicitly show that quantum entanglement can be used for guiding the learning of data for image recognition. By training the unitary MPS, our numerical experiments demonstrate that the bipartite entanglement entropy indicates the complexity of the tasks using classical computations. The single-site entanglement entropy characterizes the importance of the data to the classification problems, with which an optimization technique of the MPS architecture is proposed to largely improve the efficiency. 
	
	Our proposal can be readily applied to feature extraction, and to improving the efficiency of other learning schemes, such as those based on hierarchical TN's. The exploitation of DCT implies that quantum techniques such as TN can be combined with classical computational techniques, such as neural networks, to develop novel efficient learning algorithms. Revealing the relations to theoretical physics (e.g., quantum information) would provide a solid ground for TN machine learning, avoiding being a ``trial-and-error alchemy''.
	
	From the viewpoint of quantum computation for machine learning \cite{PanJW,Seth2013,MLscCircuit}, there are two advantages of our proposal. Firstly, the MPS we train is formed by unitaries, which has good accuracy with relatively small bond dimensions. Note that in principle, any local unitary maps or gates can be realized in quantum simulators or computers. Secondly, our proposal permits to largely reduce the size of the MPS (meaning the numbers of both qubits and quantum gates) without harming the accuracy. This significantly lowers the complexity of quantum computations, which strongly depends on the numbers of the qubits and gates. The reduced number of qubits is only around $50 \sim 100$, which is within the access of the state-of-the-art quantum computers. The low demands on the bond dimensions and, particularly, on the size, permit to simulate machine learning tasks by quantum simulations or quantum computations in the near future.
	
	\section*{acknowledgement}
	S.J.R. thanks Gang Su, Lei Wang, Ding Liu, Cheng Peng, Zheng-Zhi Sun, Ivan Glasser, and Peter Wittek for stimulating discussions. Y.H.L thanks Naichao Hu for helpful suggestions of writing the manuscript. This work was supported by ERC AdG OSYRIS (ERC-2013-AdG Grant No. 339106), Spanish Ministry MINECO (National Plan 15 Grant: FISICATEAMO No. FIS2016-79508-P, SEVERO OCHOA No. SEV-2015-0522), Generalitat de Catalunya (AGAUR Grant No. 2017 SGR 1341 and CERCA/Program), Fundaci\'o Privada Cellex, EU FETPRO QUIC (H2020-FETPROACT-2014 No. 641122), the National Science Centre, and Poland-Symfonia Grant No. 2016/20/W/ST4/00314. S.J.R. was supported by Fundaci\'o Catalunya - La Pedrera . Ignacio Cirac Program Chair. Y.H.L is supported by National Program For Top-notch Undergraduate in Basic Science under Grant No. 03100-31911002 from the Ministry of Education of P.R. China. X.Z. is supported by the National Natural Science Foundation of China (No. 11404413), the Natural Science Foundation of Guangdong Province (No. 2015A030313188), and the Guangdong Science and Technology Innovation Youth Talent Program (Grant No. 2016TQ03X688).
	
	
	\bibliographystyle{apsrev} 
	

	
	\clearpage
	
	\setcounter{equation}{0}
	\setcounter{figure}{0}
	\setcounter{table}{0}
	\setcounter{page}{1}
	\makeatletter
	\renewcommand{\theequation}{A\arabic{equation}}
	\renewcommand{\thefigure}{A\arabic{figure}}
	\renewcommand{\thetable}{A\arabic{table}}
	
	\begin{widetext}
		\section*{Appendix A: some details of the training algorithm}
		
		We introduce several tricks to speed up the training procedure. Firstly, we evolve the environment tensors $\mathbf{E}^{[l]}$ to avoid putting too many training samples in one single iteration. Specifically speaking, we only randomly select a small number of samples (say $1000$) and compute the corresponding environment tensor $\tilde{\mathbf{E}}^{[l]}$. Then we update $\mathbf{E}^{[l]} \leftarrow \mathbf{E}^{[l]} + \delta \tilde{\mathbf{E}}^{[l]}$ with $\delta$ a small constant. $\mathbf{E}^{[l]}$ is the total environment tensor and can be initialized as the $\tilde{\mathbf{E}}$ obtained in the first iteration. Then we use SVD of the total environment tensor $\mathbf{E}^{[l]} = \mathbf{U} \mathbf{\Lambda} \mathbf{V}^{\text{T}}$ to update the tensor as $\mathbf{A}^{[l]} \leftarrow \mathbf{V} \mathbf{U}^{\text{T}}$. We find this harms little the accuracy but can largely save the computational time and memory. Our simulation also shows high accuracy and fast convergence with $\delta=1$. The difference between large and small $\delta$ is the stability under certain extreme conditions, such as training with very small bond dimensions.
		
		Secondly, we restore all the intermediate vectors during the contraction process to avoid repetitive computations. This trades the computational time by memory, and do no harm to the accuracy. 
		
		
		Thirdly, we take advantage of the unitary property of the MPS. The original cost function should be the negative log-likelihood (NLL), which reads
		\begin{equation}
		\begin{aligned}
		f^{CE} = \ln \text{Tr}(\hat{\mathbf{\Psi}} \hat{\mathbf{\Psi}}^{\dagger}) -  \frac{1}{N\text{Tr}(\hat{\mathbf{\Psi}} \hat{\mathbf{\Psi}}^{\dagger})} \sum_{n=1}^N \ln( \sum_{bs_1 \cdots s_L}B^{[n]}_{b} \hat{\Psi}_{b,s_1 \cdots s_L} \prod_{l=1}^L v^{[n,l]}_{s_l}),
		\end{aligned}
		\end{equation}
		with $N$ the total number of images. Considering $\text{Tr}(\hat{\Psi} \hat{\Psi}^{\dagger})$ as a constant according to the orthogonal condition, one has
		\begin{equation}
		\begin{aligned}
		\tilde{\mathbf{E}}^{[l]} = \frac{\partial f^{CE}}{\partial \mathbf{A}^{[l]}} = \frac{1}{N} \sum_{n=1}^N  \frac{\mathbf{\mathcal{E}}^{[n,l]}} {\sum_{bs_1 \cdots s_L} B^{[n]}_{b} \hat{\Psi}_{b,s_1 \cdots s_L} \prod_{l=1}^L v^{[n,l]}_{s_l}},
		\end{aligned}
		\end{equation}
		with $\mathbf{\mathcal{E}}^{[n,l]}$ the environment tensor for the $n$-th sample without normalization. More investigations are to be done to further understand the techniques explained above [Zheng-Zhi Sun \textit{et al}, in preparation]. 
		
		
		For the feature map, the standard one maps a pixel $\theta$ satisfying $0 \leq \theta \leq 1$ to a normalized vector $\mathbf{v}$ which ranges from $[1,0]$ (spin up) to $[0,1]$ (spin down). When the feature map is fixed, the range of $\theta$ (with $0 \leq \theta \leq \tilde{\theta}$) changes with the range of $\mathbf{v}$ (from  $[1,0]$ to a canted spin state $[\cos \alpha, \sin \alpha]$), and \textit{vice versa}. It is obvious that $\sin \alpha = \tilde{\theta}$. Meanwhile, we find that by controlling $\tilde{\theta}$, accuracy can change. Without DCT, we take $0 \leq \theta \leq 1$ and $\alpha = \pi/4$, which gives relatively high precision and stability. With DCT, the signs and the maximum/minimum of the ``pixels'' (also denoted by $\theta$) of each image are not fixed. The accuracy and stability are the highest with $-1 \leq \theta \leq 1$ and $\alpha = 2\pi$. This is because with DCT, most values are quite small, which requires a relatively large $\alpha$.
		
		We shall stress that our proposal of entanglement-based architecture is independent on the algorithms or tricks for optimizing the MPS (or other TN's). Once the algorithm is chosen, our proposal can be utilized to reveal the ``quantum'' features of the machine learning tasks and improve the efficiency of the training.
		
		\section*{Appendix B: precision of the two-class classifiers on the test dataset}
		
		In Table \ref{precision2}, we show the accuracy on the test dataset for all the two-class classifiers trained by the frequency data. We take physical bond dimension $d=2$ and the virtual bond dimension $\chi=16$. In each iteration, we  feed $1000$ samples randomly picked from the two classes. 
		
		\begin{table}[tbp]
			\centering
			\begin{tabular}{ c||c|c|c|c|c|c|c|c|c } 
				\hline
				--& 1 & 2 &3&4&5&6&7&8&9 \\
				\hline\hline
				0 & 0.9981 & 0.9896& 0.9965& 0.9975& 0.9899& 0.9897& 0.9960& 0.9913& 0.9925  \\ 
				\hline
				1 & - & 0.9949& 0.9967& 0.9967& 0.9956& 0.9962& 0.9884& 0.9934& 0.9953 \\
				\hline
				2& -& -& 0.9838& 0.9940& 0.9922& 0.9940& 0.9811& 0.9855& 0.9912\\
				\hline
				3& -& -& -& 0.9980& 0.9721& 0.9995& 0.9858& 0.9738& 0.9827\\
				\hline
				4& -& -& -& -& 0.9973& 0.9871& 0.9896& 0.9939& 0.9653\\
				\hline
				5& -& -& -& -& -& 0.9881& 0.9958& 0.9845& 0.9842\\
				\hline
				6& -& -& -& -& -& -& 0.9975& 0.9948& 0.9975\\
				\hline
				7& -& -& -& -& -& -& -& 0.9885& 0.9637\\
				\hline
				8& -&-& -& -& -& -& -& -& 0.9808\\
				\hline
				\hline
			\end{tabular}
			\caption{Precision of the two-class classifiers trained by frequency data.The virtual bond dimension is $\chi=16$, with $D=2$ and $d=2$.}
			\label{precision2}
		\end{table}
		
		For comparison, the accuracy obtained from the real-space data is shown in Table \ref{precision_real}. In general, the accuracy from the frequency data is generally at the save level with that from the real-space data. This is expected since the DCT gives a unitary transformation on the data.
		
		\begin{table}[tbp]
			\centering
			\begin{tabular}{ c||c|c|c|c|c|c|c|c|c } 
				\hline
				--& 1 & 2 &3&4&5&6&7&8&9 \\
				\hline\hline
				0 & 0.9980 & 0.9950& 0.9990& 0.9970& 0.9980& 0.9890& 0.9940& 0.9960& 0.9960  \\ 
				\hline
				1 & - & 0.9950& 0.9960& 0.9990& 0.9990& 0.9960& 0.9970& 0.9980& 0.9990 \\
				\hline
				2& -& -& 0.9850& 0.9980& 0.9970& 0.9970& 0.9900& 0.9950& 0.9990\\
				\hline
				3& -& -& -& 1& 0.9930& 1& 0.9930& 0.9920& 0.9950\\
				\hline
				4& -& -& -& -& 1& 0.9970& 0.9980& 0.9970& 0.9940\\
				\hline
				5& -& -& -& -& -& 0.9930& 0.9990& 0.9910& 0.9890\\
				\hline
				6& -& -& -& -& -& -& 1& 0.9980& 0.9970\\
				\hline
				7& -& -& -& -& -& -& -& 0.9940& 0.9920\\
				\hline
				8& -&-& -& -& -& -& -& -& 0.9830\\
				\hline
				\hline
			\end{tabular}
			\caption{Precision of the two-class classifiers trained by real-space data.The virtual bond dimension is $\chi=16$, with $D=2$ and $d=2$.}
			\label{precision_real}
		\end{table}
		

		\section*{Appendix C: SEE for real-space MPS classifiers}
		
		For the real-space MPS classifiers, SEE can characterize the importance on different sites of the images. This can be viewed clearly from the SEE distribution viewed in 2D plane (Fig. \ref{fig-S1}). We see that the SEE distribution captures the main features of the two classes of images that are to be classified. With the path optimization, the extracted features of the images are stored not only in the SEE, but also in the path of the MPS (i.e., how the MPS covers the 2D image).
		
		Besides, we notice that with the real-space data, SEE is zero along the edges of about 4-pixel width, corresponding to the blank edges of most of the images in the MNIST images. This serves as another proof that SEE characterizes the importance of the data provided on different sites.
		
		\begin{figure}[H]
			\centering
			\subfloat{	\includegraphics[width=0.16\linewidth]{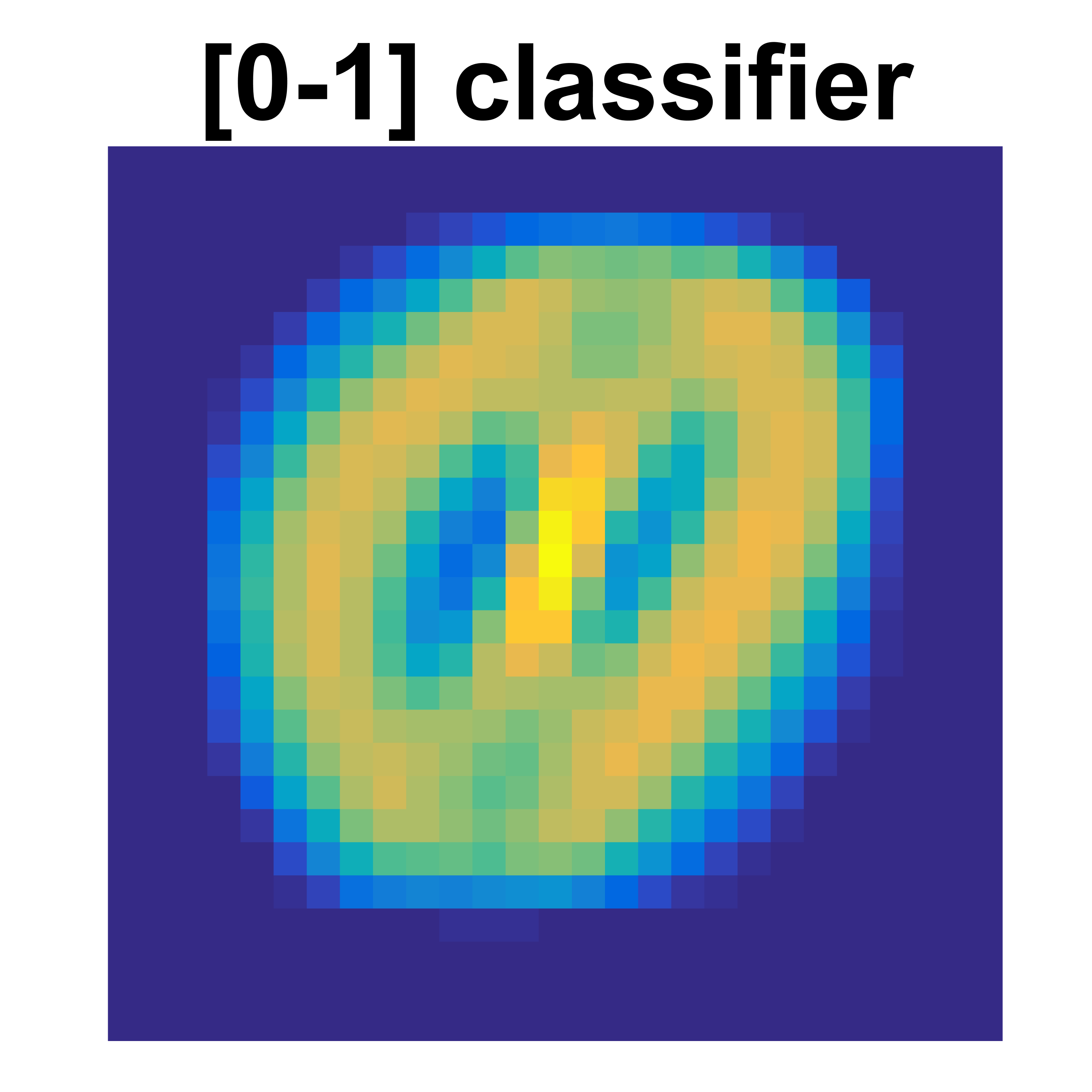}}
			\subfloat{\includegraphics[width=0.16\linewidth]{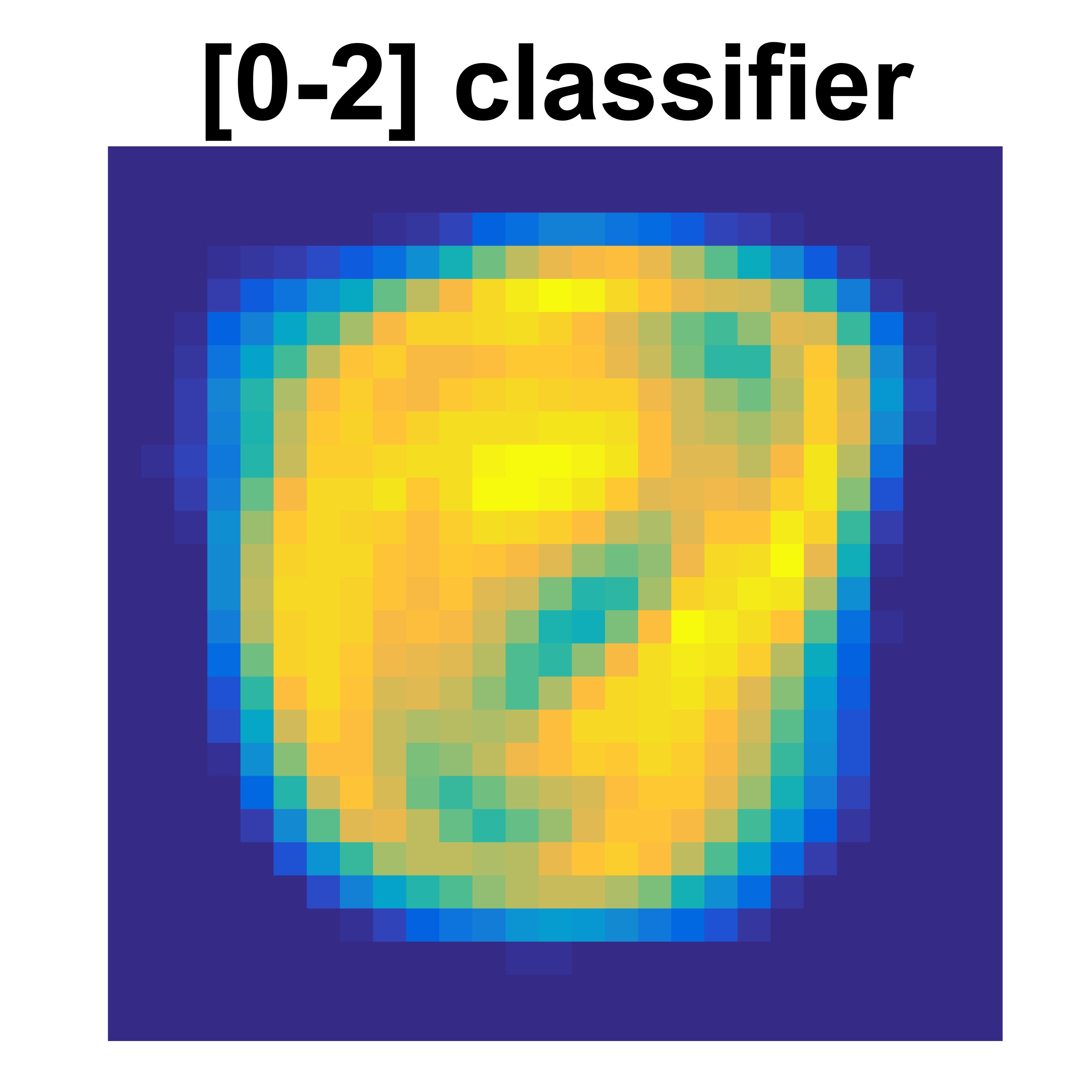}}
			\subfloat{	\includegraphics[width=0.16\linewidth]{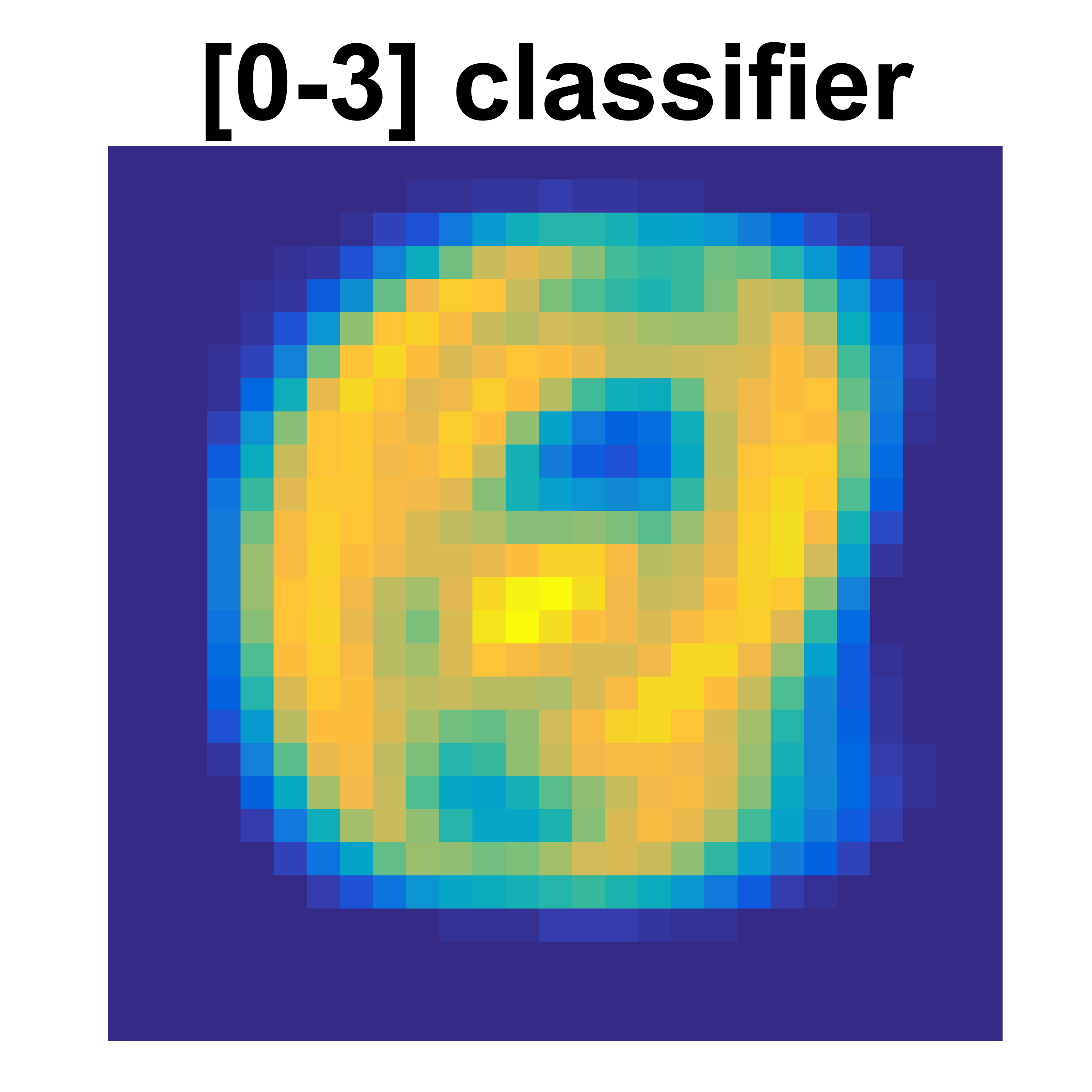}}
			\subfloat{	\includegraphics[width=0.16\linewidth]{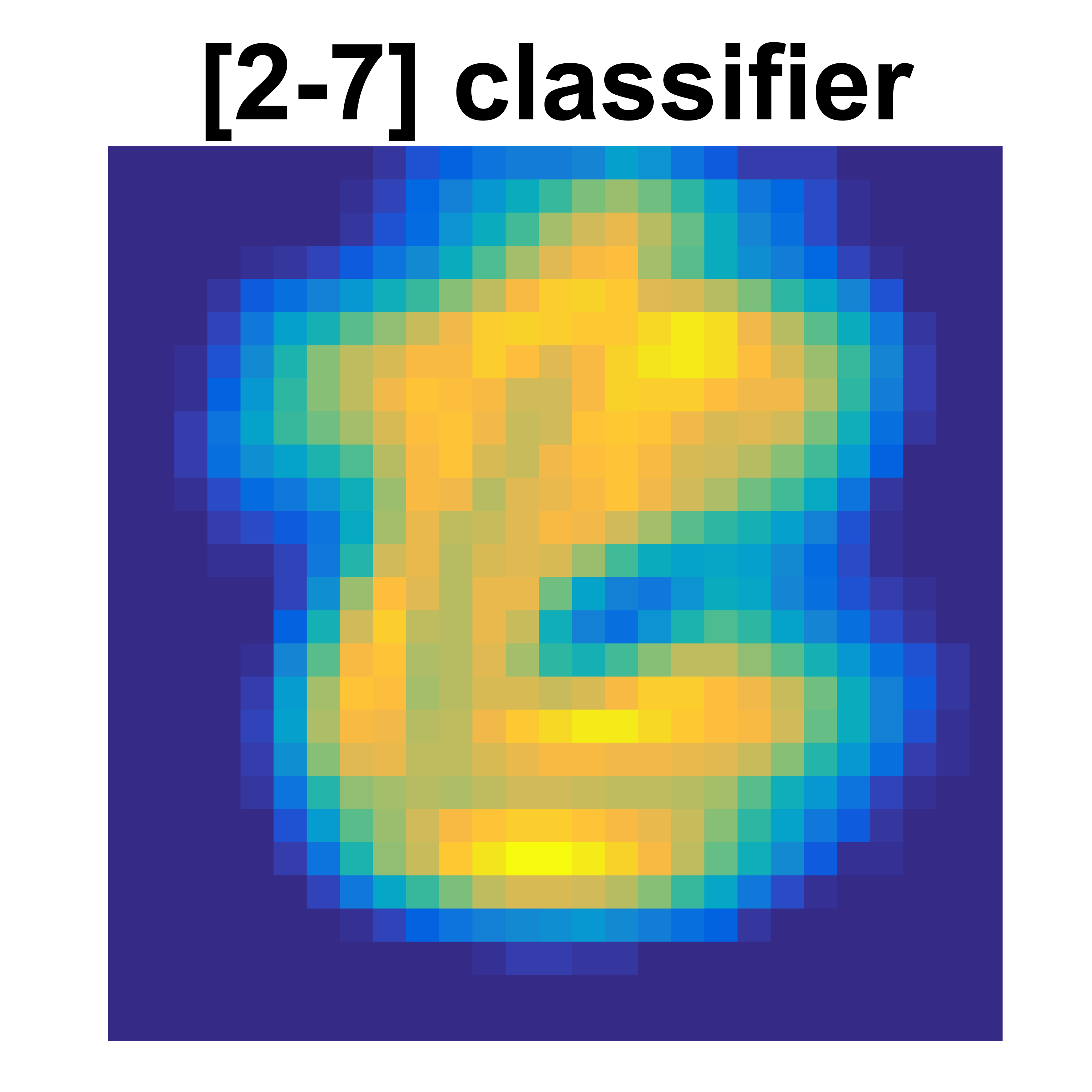}}
			\subfloat{	\includegraphics[width=0.16\linewidth]{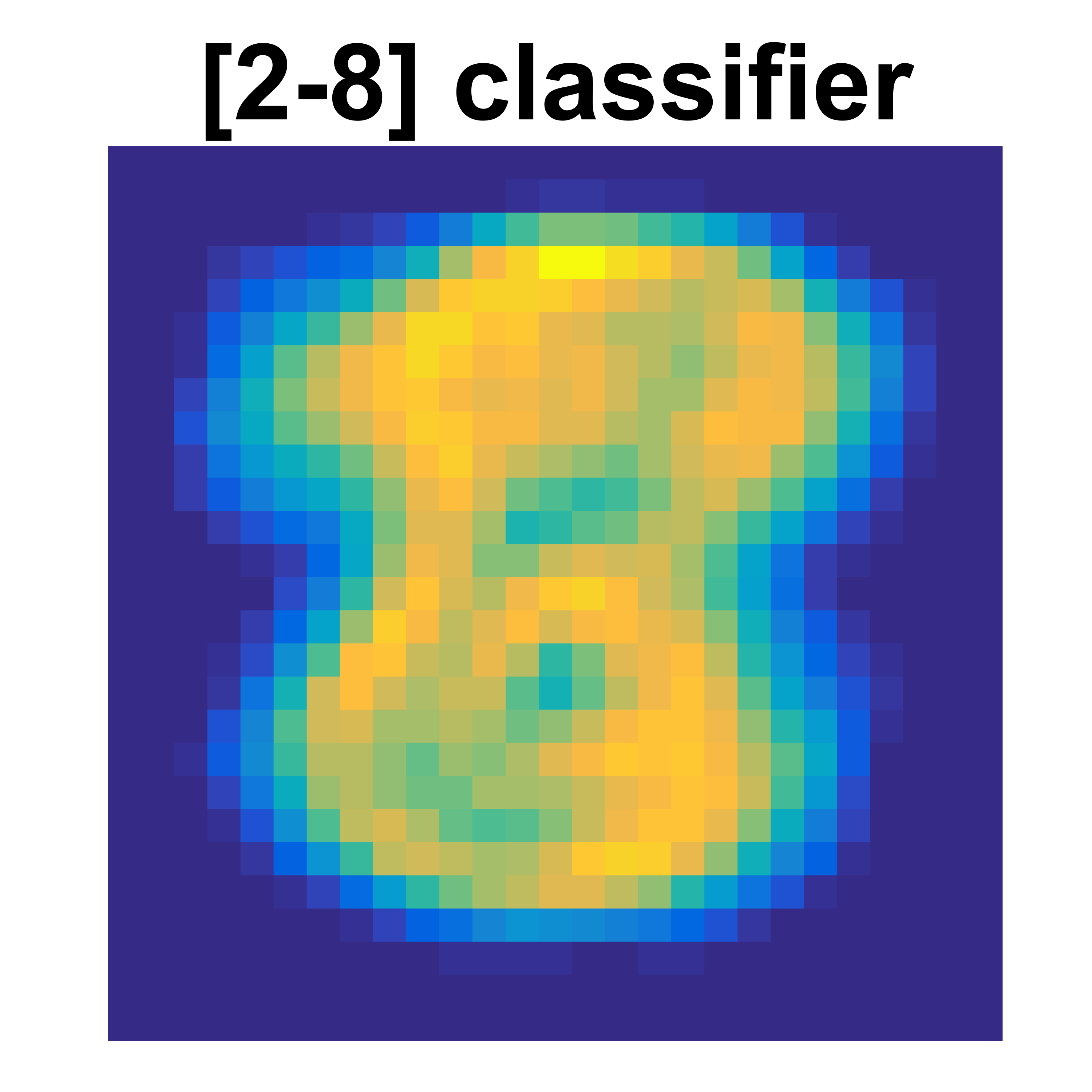}}
			\subfloat{	\includegraphics[width=0.038\linewidth]{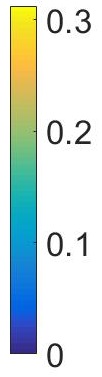}}
			\caption{SEE for five real-space classifiers, from left to right: $[0,1]$, $[0,2]$, $[0,3]$, $[2,7]$, and $[2,8]$ classifiers. One can see that the SEE can capture the features of the images.}
			\label{fig-S1}
		\end{figure}
		
	\end{widetext} 
    
\end{document}